\documentclass[runningheads]{llncs}

\usepackage[mobile]{eccv}

\usepackage{eccvabbrv}

\usepackage{graphicx}
\usepackage{booktabs}
\usepackage{makecell}
\usepackage{algorithm}
\usepackage{algpseudocode}

\usepackage[accsupp]{axessibility}  %

\usepackage{hyperref}

\usepackage{orcidlink}

\begin{document}

\title{Improved Immiscible Diffusion: Accelerate Diffusion Training by Reducing Its Miscibility} 

\titlerunning{Improved Immiscible Diffusion}

\author{Yiheng Li\inst{1}\orcidlink{0000-0002-2706-7147} \and
Feng Liang\inst{2}\orcidlink{0000-0002-1619-9355} \and
Dan Kondratyuk\inst{3} \and
Masayoshi Tomizuka\inst{1}\orcidlink{0000-0003-0206-6639} \and
Kurt Keutzer\inst{1}\orcidlink{0000-0003-3868-8501} \and
Chenfeng Xu\inst{2}\thanks{Corresponding author.}\orcidlink{0000-0002-4941-6985}}

\authorrunning{Y. Li et al.}

\institute{UC Berkeley, Berkeley CA 94720, USA \and
UT Austin, Austin TX 78712, USA \and
Rekursiv.ai, Mountain View CA, 94041 \\
\email{\{yhli, tomizuka, keutzer\}@berkeley.edu, \{jeffliang, cxu\}@utexas.edu, dan@rekursiv.ai}}

\maketitle

\begin{figure}[htbp]
  \vspace{-2em}
  \centering
  \includegraphics[width=1.00\linewidth]{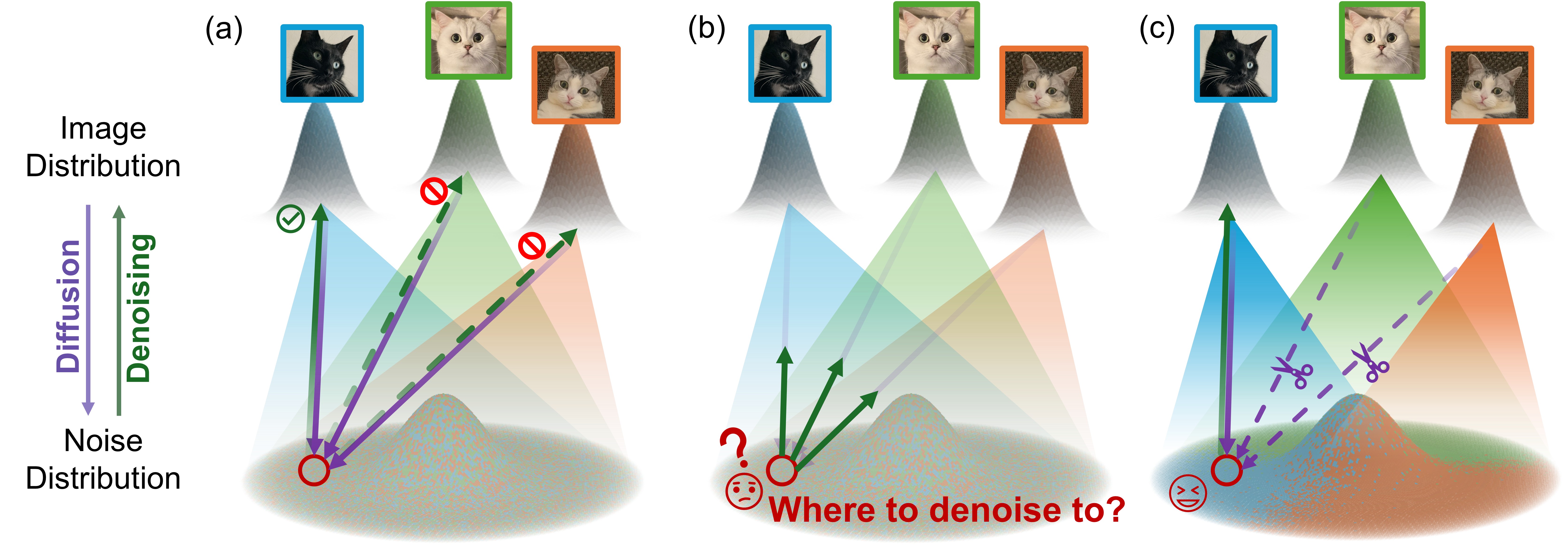} %
  \vspace{-1em}
  \caption{\textbf{Improved Immiscible Diffusion. (a)} While vanilla diffusion trajectories (flows) are mixed (miscible), \textit{each noise point is stably correlated to a specific generated image}, making many crossed diffusion trajectories irreversible. \textbf{(b)} The irreversible trajectories confuse the denoising process. \textbf{(c)} We extend immiscible diffusion to cut mixed (miscible) diffusion trajectories during diffusion training for accelerating it.}
  \label{fig:teaser}
  \vspace{-4em}
\end{figure}

\begin{abstract}
  The substantial training cost of diffusion models hinders their deployment. Immiscible Diffusion~\cite{immiscible_diffusion_nips} showed that reducing the mixing of images' diffusion destination in the noise space via linear assignment can accelerate diffusion training. However, concerns regarding limited image diversity and exploding execution times under large batch sizes limit its feasibility for large-scale training. In this work, we start from thoroughly tackling these limitations: For the image diversity, we demonstrate the bijective nature of the denoising process of vanilla diffusion, underlying that being immiscible cannot hurt the diversity. Moving beyond the noise layer, we refine immiscible diffusion's concept to a broader miscibility reduction at any layer, which enables us to propose a new family of its implementations much more efficient to execute under high batch sizes, including K-nearest neighbor (KNN) noise selection and image scaling. Overall, the immiscible diffusion family achieves up to $4\times$ faster training across diverse models and tasks, including unconditional/conditional generation, image editing, and robotics planning. Extensive analysis shows step-by-step on how immiscibility eases denoising and improves efficiency. Besides, our analysis of immiscibility offers a novel perspective on how optimal transport (OT) enhances diffusion training. By identifying trajectory miscibility as a fundamental bottleneck, we believe this work establishes a potentially new direction for future research in high-efficiency diffusion training.
  \vspace{-1em}
\end{abstract}

\section{Introduction}
\vspace{-6pt}

Training diffusion-based models is costly and time-consuming, and such training costs are continuously growing. For example, Stable Diffusion V1.1 \cite{stable_diffusion_paper} was trained for more than $24$ days on $256$ GPUs, while its V2's training time grew to $32$ days with the same GPUs. Consequently, the training efficiency problem attracts intensive explorations from diverse aspects \cite{stable_diffusion_paper, flow_matching, Multisample_Flow_Matching, OTCFM}. 

Recent works~\cite{Multisample_Flow_Matching, OTCFM} found that applying batch-wise optimal transport (OT) to pair the images and the noises can boost the training efficiency of flow matching~\cite{flow_matching} by shortening flow trajectories and lowering variance in denoising. However, immiscible diffusion \cite{immiscible_diffusion_nips} shows that such approximate OT reduces the image-noise distance by only $\approx2\%$, and multisample flow matching \cite{Multisample_Flow_Matching} shows that the standard deviation (std) of the denoising function reduces only $\approx4\%$. On the other hand, immiscible diffusion \cite{immiscible_diffusion_nips} assigns each image to a relatively separated noise area to boost the training efficiency, attributing such performance improvements to better denoising performances in noisy layers. However, its implementation is limited to image-noise linear assignment, which not only reduces image-noise distance for just $\approx2\%$ as well, but also has a time complexity of $O(n^3)$, questioning its efficiency optimality. Additionally, the concept of immiscible diffusion itself raises questions on the diversity of its generated images, as it stops images from being diffused to some far-away noise areas.

In this work, we first refine immiscible diffusion to a broader concept: a diffusion process with reduced mixtures of diffusion trajectories (flows) from different images. We find that generated images are stably correlated to their noise origins, which makes the mixing of diffusion trajectories unnecessary. This resolves the diversity concerns of immiscible diffusion. We then explore immiscible diffusion's benefits step-by-step, demonstrating how the miscibility reduction eventually leads to boosts on diffusion models. With the refined definition, we accordingly offer improved immiscible diffusion implementations including KNN and image scaling, which satisfy the improved concept but either do not qualify for an assignment or even involve no image-noise pairings. 

Extensive experiments are performed to examine the performance of the new immiscible diffusion family, where we observe consistent training efficiency boosts on various baseline models including consistency models~\cite{consistency_model}, flow matching~\cite{flow_matching} and DDIM~\cite{DDIM}. Further experiments see similar boosts across diverse image generation tasks including unconditional and conditional ones, different stages (training and fine-tuning), and various image datasets such as CIFAR-10~\cite{CIFAR10}, ImageNet~\cite{IMAGENET} and MSCOCO~\cite{mscoco}. Moreover, the immiscible diffusion family is subsequently extended to tasks outside image generation such as image in-painting and out-painting and robotics planning~\cite{diffusion_policy}, where the coherent performance enhancements support its robustness. Thorough discussion is provided to distinct immiscible diffusion to other training efficiency improvement methods, and to compare between the implementations. Our contributions are summarized as follows,

\begin{itemize}
    \item We extend immiscible diffusion~\cite{immiscible_diffusion_nips} to an implementation-agnostic concept, characterized by reduced mixture (miscibility) of diffusion trajectories from different images. Our experiments show that generated images are stably correlated with their noise origins, relieving concerns on immiscible diffusion's generation diversity. Systematic feature analysis clarifies how immiscible diffusion enhances diffusion training.

    \item Based on the improved immiscible diffusion concept, we design a family of implementations, including the KNN, image scaling. These methods are more efficient to linear assignment, and feature analysis shows that they both effectively reduce the miscibility of diffusion trajectories.

    \item The immiscible diffusion family is applied to various image generation tasks, including unconditional and conditional image generation training and fine-tuning, and across diverse datasets and baseline methods. Unanimous effectiveness of immiscible diffusion is observed, and additional experiments extend its benefits to applications such as image editing and robotics planning. The miscibility problem we established points out a potential direction for future research in efficient diffusion training.
    
\end{itemize}

\vspace{-9pt}
\section{Related Works}
\vspace{-6pt}
\subsection{Training Efficiency of Diffusion-based Models}

As training efficiency limits the large-scale deployment of diffusion-based models, actions are taken to trigger diverse abundant parts inside them. For the feature dimensions, Stable Diffusion~\cite{stable_diffusion_paper} shrinks image sizes with VAE~\cite{VAE}, downsizing the image dimension by 16-64 times. \cite{patch_diffusion} introduces a novel patch strategy to control the ease of diffusion training, achieving both training and data efficiency. For the noise space, \cite{blue_noise} modifies the noise used in diffusion models to boost the performance. For training dynamics, \cite{karras_training_dynamics} discovers that the magnitude of activation and the magnitude of neural weights can impact the diffusion's training speed. \cite{improved_consistency_model} demonstrates the importance of training dynamics on training efficiency, including the usage of exponential moving average (EMA), training loss and the noise schedule. \cite{acmmm_train_unnoisy_steps, SpeeD} notices that denoising some noisy steps helps little, so focusing more on other steps can improve training efficiency. However, these works hardly alter the diffusion trajectory, therefore cannot solve the miscibility problem immiscible diffusion aims to tackle.
\vspace{-6pt}
\subsection{Diffusion Paths and Training Efficiency}

Recent works redesign diffusion trajectories for faster training. Notably, based on the rectified flow~\cite{rect_flow}, \cite{instaflow} improves training efficiency by making diffusion trajectories deterministic and straight. \cite{flow_matching} further straightens the diffusion trajectory by making the image-noise mixture linear. Probing more deeply, several studies began exploring alternatives to mapping each image to the full noise space. \cite{curvature_non_gaussian} pointed out the curvature problem in the ODE paths caused by the collapse of the denoising trajectories pointing to the average direction. However, they replace the noise sampling with a VAE encoder-style structure to eliminate such curvatures, which destroys the strict Gaussian of the noise. In order to make diffusion trajectory paths even straighter and shorter, \cite{Multisample_Flow_Matching, OTCFM} applies batch-wise OT to assign noise to closer images before performing flow matching, followed up by a few further improvements on them \cite{xing2023exploringstraightertrajectoriesflow, deng2024variationalschrodingerdiffusionmodels, chemseddine2024conditionalwassersteindistancesapplications, issenhuth2024improvingconsistencymodelsgeneratorinduced, ofm}. However, \cite{immiscible_diffusion_nips} shows that the OT only reduces average image-noise distance by $\approx2\%$, and \cite{Multisample_Flow_Matching} claims that the standard deviation of denoising reduces only $\approx4\%$ after applying OT, which doubts the contributions of OT in the training efficiency boosts. \cite{xu2023stable} trains diffusion models to denoise towards the weighted average of a large image batch, to avoid large denoising variance. However, the weighted average calculations introduced place additional costs to the training. More recently, immiscible diffusion~\cite{immiscible_diffusion_nips} assigns noise to some preferred noise areas, aiming to avoid the difficulty of denoising in noisy layers. However, its implementation of batch-wise linear assignment is still similar to OT, so its theory needs to be further justified versus OT.
\vspace{-6pt}
\subsection{Image-Noise Correlation in Diffusion Models}

While most diffusion-based models diffuse each image to the whole noise space, some diffusion inversion works indicate that learned diffusion model's inversion does not follow this. \cite{there_and_back} indicates that the diffusion and denoising is not symmetric, and \cite{diff_diffusion_same_img_same_noise} demonstrates that nearly the same image would be generated with the same noise but diverse diffusion models. These imply that generated images and their noise origins are somehow correlated. \cite{image_noise_similarity} further illustrates some similarity between the generated images and their noise origin, so as \cite{workshop_diff_method_similarity}. Quantitatively, \cite{inversion_ot, nips23_ot_dataset_noise, diffusion_by_learning_ot} suggests that DDPM's inversion is an $L_2$ OT process. However, the correlation strength between generated images and their noise origin, \textit{i.e.} how much perturbation can such correlation resist, was not thoroughly discussed. Our work proves a \textbf{stable} relation exists between the generated images and their noise origins, which supports the generation diversity of immiscible diffusion.

\vspace{-6pt}
\section{Preliminaries}

For vanilla diffusion models, during diffusion training, we add random Gaussian noise to images,

\begin{equation}
    x_t = \sqrt{\alpha_t} x_0+\sqrt{1-\alpha_t}\epsilon\space, \space\epsilon\in N(0,I)
\end{equation}

where $x_0$ represents the image, $\epsilon$ is the added Gaussian noise, and $\alpha_t$ is the noise schedule for each diffusion timestep $t\in[0,T]$ controlling the mixing weights of $x_0$ and $\epsilon$. $x_t$ is the noisy image we get, and we train the model to predict $\epsilon_\theta(x_t, t)$ with $\epsilon$ as supervision.

However, such a method can mix different image's diffusion paths at a same point in the image space. For example, we have two images $x_{0,1}$ and $x_{0,2}$. Since $\epsilon$ is randomly sampled, for some $t$'s (especailly when $t\to T$) there exists $\epsilon_1$ and $\epsilon_2$, so that,
\vspace{-3pt}
\begin{equation}
    x_t = \sqrt{\alpha_t} x_{0,1}+\sqrt{1-\alpha_t}\epsilon_{1} = \sqrt{\alpha_t} x_{0,2}+\sqrt{1-\alpha_t}\epsilon_{2}
\end{equation}

We call this \textit{miscible diffusion}, where different $x_0$'s diffusion path mix at a same $x_t$. 

Such miscible diffusion can cause difficulties in denoising. For a specific point $x_t$, the denoising function $\epsilon_\theta(x_t, t)$ can only provide one single prediction. However, it is trained to predict both $\epsilon_{1}$ and $\epsilon_{2}$, so it ends up predicting

\begin{equation}
    \epsilon_\theta(x_t, t) \to \frac{1}{2}\Sigma_{i=1}^2\epsilon_{i}
\end{equation}

which cannot be used to denoise to either $x_0$'s. Most significantly, when $t \to T$, the noise schedule $\alpha_t \to 0$, so $x_t \to \epsilon$. In this case, any specific point $x_t$ can be reached by the diffusion paths originated from all images $x_0$'s. Therefore, the denoising function will converge to
\vspace{-3pt}
\begin{equation}
\begin{aligned}
    \epsilon_\theta(x_t, t) \to & \frac{1}{N}\Sigma_{i=1}^N\epsilon_{i} \\ & = \frac{1}{N\sqrt{1-\alpha_t}}\Sigma_{i=1}^N(x_t-\sqrt{\alpha_t}x_{0,i}) \\ & = \frac{1}{\sqrt{1-\alpha_t}}(x_t-\sqrt{\alpha_t}\bar{x_{0}})
\end{aligned}
\end{equation}

where $N$ is the size of the image dataset. This training goal converges to constant regardless of $x_0$, as $x_0 \to const$ when $N \to +\infty$, which does not help denoise to any specific image.

Therefore, based on \cite{immiscible_diffusion_nips}'s theory, we extend immiscible diffusion, which now represents a set of methods to avoid miscible diffusion, i.e. avoid the mixing of images' diffusion paths discussed above, during diffusion training. A straightforward method to achieve so is assign $x_t$ to closest $x_0$ in terms of pixel values, by minimizing the image-noise $L2$ distance $\|\epsilon-x_0\|_2$. In this work, we introduce two example implementations based on this in Section \ref{sec:linear_assignment}. Note that while these methods can achieve immiscible diffusion, they are not the only methods. Another simple yet effective demonstration for immiscible diffusion is the image scaling method detailed in Section~\ref{sec:methods_factoring}.

\begin{figure}[ht]
    \centering
    \includegraphics[width=1.00\linewidth]{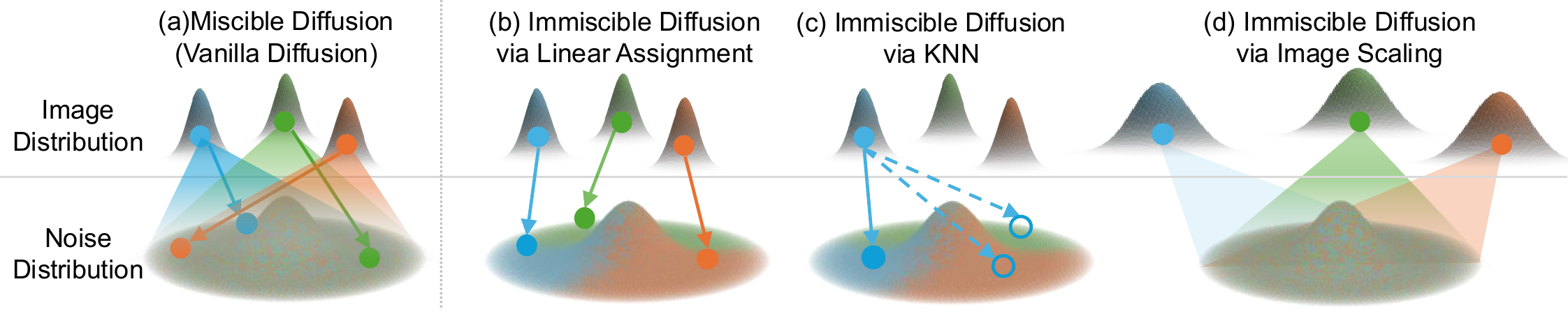}
    \vspace{-6pt}
    \caption{\textbf{Implementations of Immiscible Diffusion. (a) Miscible Diffusion} pairs the batch of images and noises randomly before adding noise to images. \textbf{(b)(c)(d) Immiscible Diffusion} tries to reduce the miscibility of diffusion by (b) $L_2$ linear assignment between the batch of images and noises and (c) sampling $k$ noises and pick the nearest one (KNN) to use. (d) scaling images by multiplying their pixel values with a constant  $>1$, which reduces overlaps between diffuse-able areas of different images.}
    \vspace{-6pt}
    \label{fig:immiscible_implementations}
\end{figure}

\vspace{-3pt}

\section{Improved Immiscible Diffusion}
\vspace{-6pt}

\cite{immiscible_diffusion_nips} borrows immiscible diffusion, a physics concept describing solutes which cannot mix homogeneously, to assign images with near noise points via batch-wise linear assignment. While this accelerates diffusion training, the linear assignment is weak with only $\approx2\%$ image-noise distance drop afterwards, and its computational complexity is $O(n^3)$, which scales up quickly with the batch size.

Therefore, to improve immiscible diffusion, we need to break the limit of using linear assignments. However, involving image-noise correlation generally triggers concerns on the diversity of generated images, as images are not diffused image to uncorrelated noises. Consequently, we need to take a deeper look into the generation diversity, to justify building correlations between images and noises.

\vspace{-6pt}
\subsection{Image-noise Correlation in Denoising}
\label{sec:immiscible_denoising}

\begin{figure}
    \vspace{-24pt}
    \centering
    \includegraphics[width=0.80\linewidth]{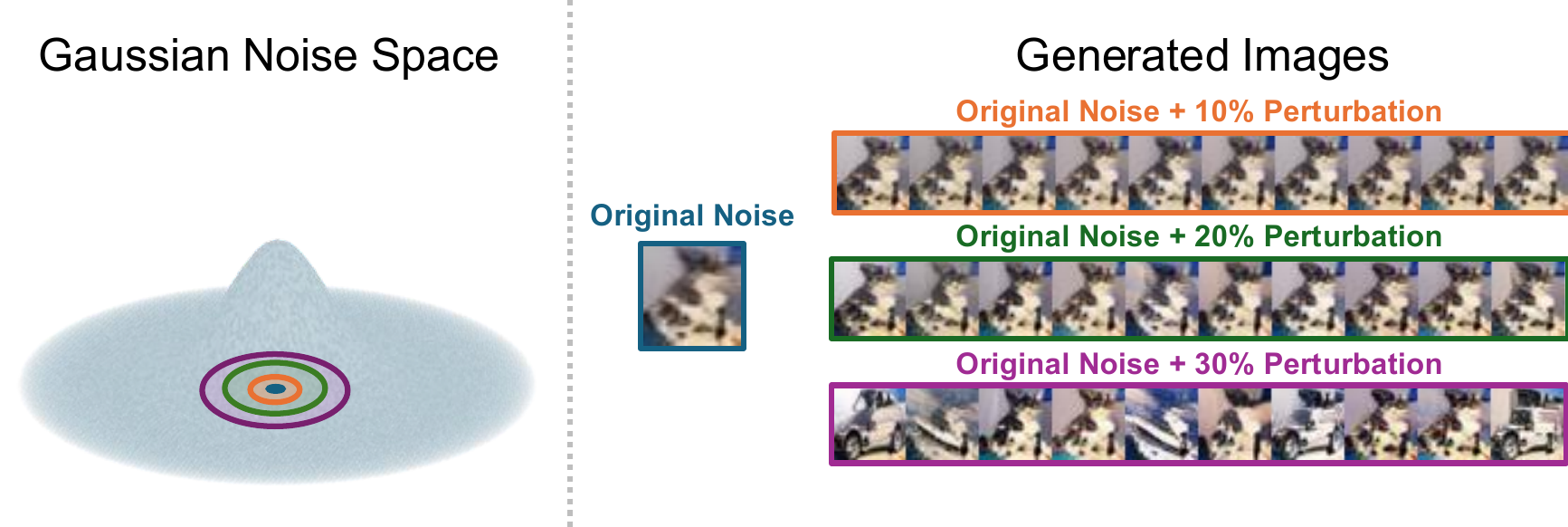}
    \vspace{-9pt}
    \caption{\textbf{Stable correlation between generated images and its noise origins.} Here perturbation means another Gaussian noise added to the fixed original Gaussian noise. Note that even with 20\% perturbation, images changes are nearly unnoticeable. Only with 30\% perturbation, a image object change happens. These demonstrate stable correlation from a noise area to a specific generated image.}
    \label{fig:correlation}
    \vspace{-12pt}
\end{figure}

To evaluate the image-noise correlation during denoising, we generate images on a trained vanilla DDIM~\cite{DDIM} with a specific noise point and its ambient noise area. We first sample a specific noise point $N_{orig}$. Then we add diverse perturbations $N_{pert}$ onto it, respectively. Specifically, we sample $10$ independent Gaussian $N_{pert}$'s. The perturbed total noise can be expressed as,
\begin{equation}
    N_{tot} = N_{orig} + W \cdot N_{pert}
\end{equation}
\vspace{-1pt}
Where $W$ is the weight of the perturbation. We present images generated by the original and perturbed total noise in Figure \ref{fig:correlation}. For example, we see $N_{orig}$ generate a cat image. Surprisingly, with $W = 10\%$ on all $10$ perturbation, we see that the generated images are still all cats without noticeable differences. Further, we see that $W = 20\%$ of perturbation only results in very slight image differences (like the background difference in the $1st$ and $2nd$ images from left), while no noticeable modal changes are observed. Only with $30\%$ weight of perturbation, we observe image object changes in a few generated images. However, we can clearly find pixel-level similarities between these images despite modal differences. These results suggest that generated images are stably correlated with the sampled noise. As a result, while vanilla (miscible) diffusion models diffuse each image equally to the whole noise area, hoping to see that every image can be generated from a noise point or a small noise area, this goal might not be fully achieved, significantly weakening the motivation of miscible diffusion in generation diversity, so as the concerns on immiscible diffusion's generation diversity from a specific noise point or area. In Section \ref{sec:result_cond}, we further quantitatively show that immiscible diffusion does not negatively impact the diversity of generated images.

\vspace{-6pt}
\subsection{Step-by-step Feature-level Benefits Analysis of Immiscible Diffusion}
\vspace{-12pt}
\begin{figure*}
    \vspace{-6pt}
    \centering
    \includegraphics[width=0.90\linewidth]{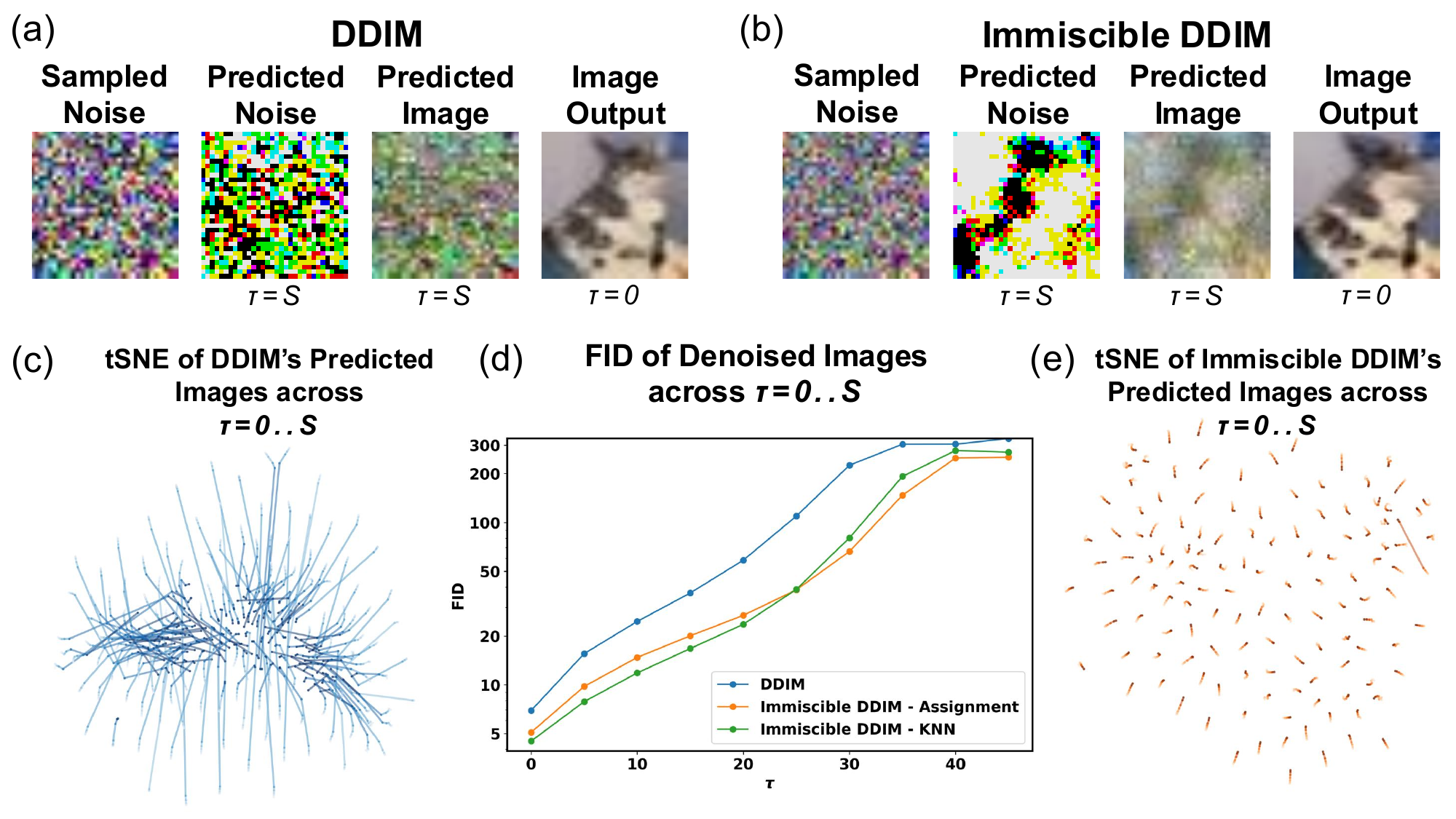}
    \vspace{-9pt}
    \caption{\textbf{Feature analysis of vanilla (miscible) and immiscible DDIM.} Referring to \cite{DDIM}, $\tau=S$ represents the layer denoising from the pure noise. We show that immiscible diffusion activates the noisiest ($\tau\to S$) layers' denoising functions by clarifying their denoising goals, as shown in the $tSNE$ of denoised images across $\tau$'s. Such activation results in FID improvements on the denoised images from large $\tau$'s, which leads to better performance and faster convergence of diffusion models.}
    \label{fig:feature_analysis}
    \vspace{-15pt}
\end{figure*}

We now dive deeper to understand more clearly on how immiscible diffusion accelerates diffusion training, in order to help to accordingly design more implementations of it. Notations used mostly follow~\cite{DDIM}, and are listed in Table~\ref{tab:notation}.

\begin{table}[t]
\centering
\small
\caption{Notations used in diffusion and denoising.}
\vspace{-6pt}
\resizebox{0.5\linewidth}{!}{
    \begin{tabular}{c|l}
    \Xhline{2\arrayrulewidth}
    Symbol & Definition \\
    \hline
    $T$ & Total \emph{diffusion} steps, $T=1000$. \\
    $S$ & Total \emph{denoising} steps, $S \in [20,50,\dots]$. \\
    $t$ & \emph{Diffusion} step, $t \in [0, T]$, $t=0$ is image. \\
    $\tau$ & \emph{Denoising} step, $\tau \in [0, S]$, $\tau=0$ is image. \\
    \Xhline{2\arrayrulewidth}
    \end{tabular}
}
\label{tab:notation}
\vspace{-12pt}
\end{table}

\cite{immiscible_diffusion_nips} suggests and mathematically proves that miscible diffusion causes difficulty in denoising at noisy layers (larger $\tau$). As shown in Figure \ref{fig:feature_analysis} (a), the noisiest denoising layer, referred as $\tau = S$ in DDIM~\cite{DDIM}, does not effectively predict the noise added. To confirm the ubiquity of such denoising difficulty, in Figure \ref{fig:feature_analysis} (c) we show the $tSNE$ of the predicted images from each denoising layer $\tau=0..S$. We collect $128$ noise points, generating images with them, and logging the predicted images ($x_{pred,0}$'s) from each denoising layer, which are computed with the layer's feature $x_\tau$ and the predicted noise $n_{pred, \tau}$ according to noise schedule $\alpha_\tau$:
\vspace{-3pt}
\begin{equation}
    x_{pred,0} = \frac{1}{\sqrt{\alpha_\tau}}x_\tau - \frac{\sqrt{1-\alpha_\tau}}{\sqrt{\alpha_\tau}}n_{pred, \tau}
\end{equation}
Points on the same line represent $x_{pred,0}$'s from the same initial noise on different $\tau$'s. Apparently, though different lines start from different noise points, they are chaotically tangled, suggesting that the same denoising goals are frequently shared between different denoising paths, which implies that the denoising difficulty commonly happens in the denoising. That is not surprising, as the vanilla DDIM takes miscible diffusion. To quantitatively express the miscibility, we stat the average $L_2$ distance between noise clusters (i.e. $10,000$ noise points) assigned to each image. The distance for vanilla DDIM is only $0.92 \pm 0.06$, proving that the diffusion process is truly miscible.

However, immiscible diffusion substantially overturns these difficulties. With the linear assignment implementation, we find that the average distance between noise clusters assigned to each image is increased significantly to $4.11 \pm 0.37$, suggesting that the noise clusters are more separate than the miscible diffusion, and thus is more immiscible by the definition. Figure \ref{fig:feature_analysis} (b) illustrates that under immiscible diffusion, even the noisiest layer $\tau = S$ can effectively predict the noise and can denoise the image, and Figure \ref{fig:feature_analysis} (e) shows that with immiscible diffusion, the $tSNE$ figure shows much less intersections, implying that each noise point has its own stable denoising goal, which corresponds to the goal of easing denoising. As a result, as shown in Figure \ref{fig:feature_analysis} (d), the FID of the denoised images from noisy layers exhibit significant improvement, which finally leads to the performance and training efficiency boost of immiscible diffusion.

The analysis above clearly figures out that the reduced miscibility helps to clarify the denoising goal, easing the denoising and helping it to be effective, and finally improve the performance of the miscible layers, and the final outputs. These step-by-step benefits of immiscible diffusion help us to extract the essence of it, and to design additional implementations in the following section.

\vspace{-12pt}

\subsection{Improved Immiscible Diffusion}
\vspace{-6pt}
As the benefits of immiscible diffusion can be traced back to the reduction of miscibility in diffusion, we naturally propose that the concept of immiscible diffusion should also reflect only the reduction of miscibility in diffusion, without unnecessary bounds to image-noise pairing. Under this improved concept, we argue that such assignment is only \textit{one} way to achieve immiscible diffusion. In this work, we additionally propose two new immiscible diffusion implementations, which do not need linear assignment, or even image-noise correlations. Nevertheless, both methods exhibit excellent immiscibility, and therefore boost the training efficiency significantly.

\vspace{-12pt}
\subsubsection{Batch-wise Linear Assignment}
\label{sec:linear_assignment}

Batch-wise linear assignment in \cite{immiscible_diffusion_nips} still qualifies immiscible diffusion. As shown in Figure \ref{fig:immiscible_implementations} (b), this method performs a linear assignment \cite{linear_assignment} between the batch of images and noise points sampled. Such an assignment preferably assigns noise to nearby images while keeping the general distribution of all noises Gaussian. However, this trades off the running speed for each step when the batch size scales up, as the linear assignment is an $O(n^3)$ operation.

\vspace{-12pt}
\subsubsection{KNN Noise Selection}
\label{sec:knn_method}

To avoid the scaling-up of execution time with larger batch sizes, we provide the second implementation of immiscible diffusion, the KNN method, where we sample $k$ Gaussian noise points for each image and pick the one $L_2$-closest to the image, as illustrated in Figure \ref{fig:immiscible_implementations} (c). This method is very efficient - its execution time is only $0.2 ms$ for a batch size of $256$ and would not scale up quickly when larger batches are used, as it is an $O(n)$ operation. As shown in Table \ref{tab:exec_time}, the KNN is much faster than the linear assignment, demonstrating its efficiency in achieving immiscibility. The algorithm is shown in Algorithm \ref{alg:knn}.

\begin{table}
\vspace{-18pt}
\caption{Execution Time (ms) of Immiscible Diffusion on a single A5000 GPU. }
\vspace{-6pt}
\centering
\resizebox{0.50\textwidth}{!}
{
    \begin{tabular}{c|cccc}
        \Xhline{2\arrayrulewidth}
        Batch Size & 128 & 256 & 512 & 1024\\
        \midrule
        Linear Assignment~\cite{immiscible_diffusion_nips} & 5.4 & 6.7 & 8.8 & 22.8 \\
        KNN & 0.2 & 0.2 & 0.3 & 0.7 \\
        $t_{assign} / t_{knn}$ & 27.0x & 33.5x & 29.3x & 32.6x \\
        
        \Xhline{2\arrayrulewidth}
    \end{tabular}
}
\label{tab:exec_time}
\vspace{-24pt}
\end{table}

\begin{algorithm}
\footnotesize
\caption{Immiscible Diffusion Implementation - KNN Sampling}
\begin{algorithmic}[1]
\State \textbf{Input:} Image $x$, $k$ random noises $\{n_1, \ldots, n_k\}$, noise schedule $\alpha_t$ 
\State $n$ $\gets$ $\mathop{\arg\min}_{n_j \in \{n_1, n_2, \ldots, n_k\}} \operatorname{dist}(x, n_j)$
\State $x_t$ $\gets$ $\sqrt{\alpha_t} x + \sqrt{1 - \alpha_t} \cdot n$
\State \textbf{Output:} Diffused image batch $x_t$
\end{algorithmic}
\label{alg:knn}
\end{algorithm}
\vspace{-18pt}

While the distribution of overall noise points used in training KNN-implemented immiscible diffusion is not guaranteed to be Gaussian, as some sampled noise points are selectively dropped, we argue that such discrepancy is negligible. To prove this, we sample $50k$ Gaussian noise points in the size of CIFAR-10~\cite{CIFAR10} images, i.e. $3 \times 32 \times 32$. Comparing them with the Gaussian distribution results in a KL divergence of $48.25$. We then collect another $50k$ noise points which are \textit{selected} in the KNN Immiscible Diffusion ($k = 8$), finding that their KL divergence to Gaussian is $48.60$, which is only very slightly higher than noise points from a strict Gaussian distribution. Therefore, our KNN implementation doesn't significantly alter the Gaussian noise distribution.

\vspace{-6pt}
\subsubsection{Image Scaling}
\label{sec:methods_factoring}

Though image-noise correlation is effective in building immiscible diffusion, there are also ways to reduce miscibility without it. A typical example is image scaling, \textit{i.e.} to multiply images' all pixel values by a constant factor greater than $1$, like $2$ or $4$. Such action has no influence on the noise space. However, the $L_2$ distance between each image is farther after the scaling, making the centers of diffused areas ($t < t_{max}$) of different images farther away from each other. Considering the noise amplitude at each diffusion step $t$ is kept consistent, the diffused areas for different images at every step $t < t_{max}$ are naturally having less intersections, which constitutes immiscible diffusion. We will defer the immiscibility and performance experiments of this method to Section \ref{sec:results_factoring}.

\section{Effectiveness of Immiscible Diffusion}
\vspace{-6pt}

With the improved immiscible diffusion concept and the new and existing implementations, we perform a large series of experiments to systematically examine the benefits and the generalization ability of immiscible diffusion across various image generation tasks, models, and datasets, as well as extended tasks including image in-painting, out-painting, and robotics planning.

\vspace{-3pt}

\subsection{Experiment Setups}
\label{sec:setup}
\vspace{-3pt}
 
We implement Immiscible Diffusion on diverse diffusion-based methods, including Consistency Models \cite{consistency_model}, DDIM \cite{DDIM}, Stable Diffusion \cite{stable_diffusion_paper}, and Flow Matching \cite{flow_matching}. We train these implementations on a variety of popular datasets, including CIFAR-10 \cite{CIFAR10}, ImageNet-1k \cite{IMAGENET} and MS-COCO \cite{mscoco}. The training hyper-parameters are discussed in Section \ref{sec:spp_exp_setup} and Table \ref{tab:train_param} of the Supplementary Material. 

\vspace{-3pt}

\subsection{Unconditional Image Generation Training}
\vspace{-3pt}

\begin{figure}
  \centering
  \includegraphics[width=\textwidth]{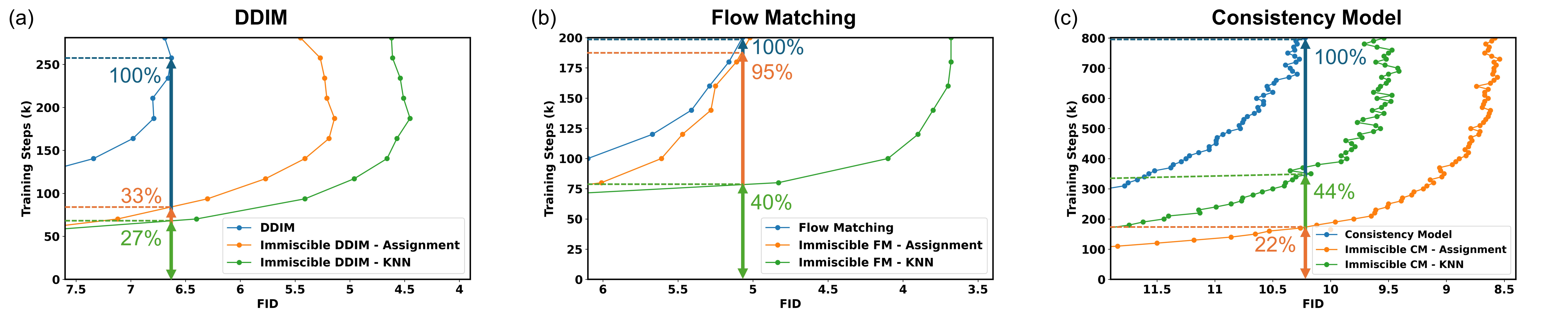}
  \vspace{-12pt}
  \caption{\textbf{Immiscible diffusion boosts training efficiency.} We show the training steps required to reach the best FID for vanilla models across \textbf{three} diverse diffusion-based architectures. Results consistently show that immiscible diffusion trains significantly faster.}
  \label{fig:uncond_fid}
  \vspace{-1.5pt}
\end{figure}

We compare the unconditional image generation training steps necessary to reach the best FID of the vanilla diffusion-based models in Figure \ref{fig:uncond_fid}. Astonishing and consistent training efficiency enhancement is observed across all diffusion-based models, despite their significant differences in diffusion trajectories (flows), denoising solvers and sampling step picking strategies. Across all experiments, we generally observe that the baseline diffusion-based models need a maximum of \textbf{2.5 to 4.5} times of training steps to achieve the same performance of their immiscible diffusion counterparts. Besides, we offer the converged FIDs in Tab.~\ref{tab:model_comparison}, confirming that immiscible models achieve better performance in most cases. These results strongly support the robust ability of immiscible diffusion in improving the training efficiency of diverse diffusion-based models. Furthermore, as shown in Tab.~\ref{tab:comp_3}, it is parallel to many popular diffusion training acceleration methods, as the improved immiscible diffusion modifies only the goal diffusion path for training.

\begin{table}
\caption{Final Performances of Immiscible Diffusion.}
\vspace{-9pt}
\centering
\resizebox{0.65\linewidth}{!}
{
    \begin{tabular}{l|ccc}
        \Xhline{\arrayrulewidth}
        Model & DDIM & Flow Matching & Consistency Models \\
        \midrule
        Vanilla & 6.63 & 4.92 & 10.22 \\
        Immiscible (KNN) & 4.45 (\textcolor{green}{-2.18}) & 3.56 (\textcolor{green}{-1.36}) & 9.16 (\textcolor{green}{-1.06}) \\
        Immiscible (Assignment) & 5.14 (\textcolor{green}{-1.49}) & 5.02 (\textcolor{red}{+0.10}) & 8.54 (\textcolor{green}{-1.68}) \\
        \Xhline{\arrayrulewidth}
    \end{tabular}
}
\label{tab:model_comparison}
\end{table}

\begin{table}
\caption{Superpositioning Immiscible Diffusion (ID) to Other Accelerative Methods.}
\vspace{-9pt}
\centering
\resizebox{0.8\linewidth}{!}
{
    \begin{tabular}{c|cccccc}
        \Xhline{\arrayrulewidth}
         & FM Base & ID & FM +~\cite{kim2024denoising} & FM +~\cite{kim2024denoising} + ID & FM+~\cite{choi2022perception} & FM +~\cite{choi2022perception} + ID \\
        \midrule
        FID \@ 100k & 6.10 & 4.28 & 6.43 & 4.75 & 5.58 & 4.26 \\
        FID \@ 200k & 5.07 & 3.68 & 5.38 & 3.93 & 5.41 & 3.86\\
        \Xhline{\arrayrulewidth}
    \end{tabular}
}
\label{tab:comp_3}
\end{table}

\subsection{Conditional Image Generation Training and Fine-tuning}
\label{sec:result_cond}
\vspace{-6pt}

\begin{figure}
  \vspace{-5pt}
  \centering
  \includegraphics[width=\textwidth]{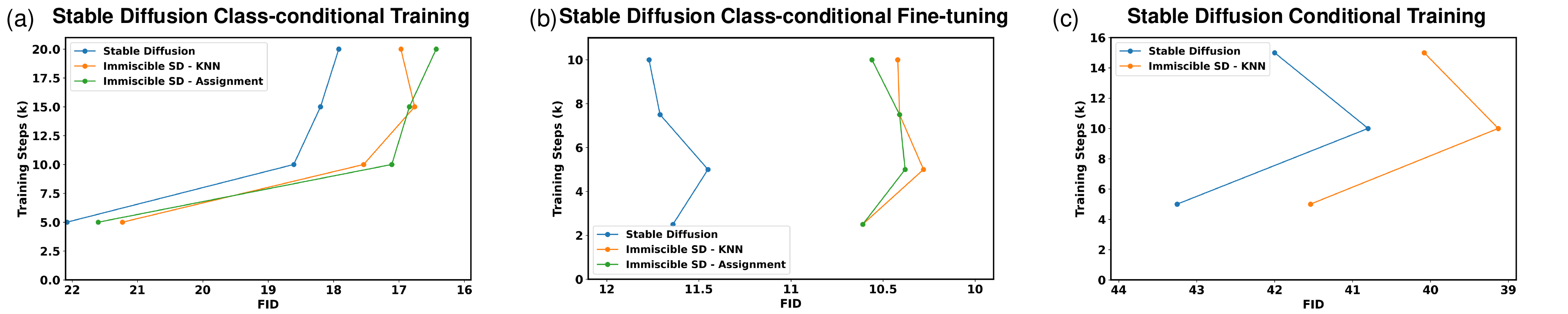}
  \vspace{-12pt}
  \caption{\textbf{Immiscible diffusion stays effective across diverse image generation tasks.} We demonstrate this with figures showing the training steps \textit{v.s.} FID of \textbf{(a)(b)} Stable Diffusion class-conditional training and fine-tuning with ImageNet, and \textbf{(c)} conditional training with MS-COCO.}
  \label{fig:cond_fid}
  \vspace{-18pt}
\end{figure}

\textbf{Class-conditional Image Generation from Scratch.} We conduct this on Stable Diffusion \cite{stable_diffusion_paper} and ImageNet-1k \cite{IMAGENET}, whose result is shown in Figure \ref{fig:cond_fid} (a). Similar to the unconditional generations, immiscible diffusion exhibits much faster training. However, since immiscible diffusion does not diffuse each image equally to the noise space, questions on whether immiscible diffusion models can still follow the prompts as good as vanilla models can be raised. Simply put, the answer is \textit{Yes, they can}. In Section \ref{sec:immiscible_denoising}, we have shown that vanilla diffusion-based models yet have strong image-noise correlations, so the diversity of generated images should not be influenced solely by the miscibility of the diffusion process. We further confirm this by evaluating the diversity of the generated images with CLIPScore \cite{clip_score}, which shows that both the immiscible and the baseline models generate images with CLIPScores of $28.55$, with a standard deviation of $0.01$ and $0.02$ respectively, indicating that Immiscible Diffusion does not hurt the image-prompt correspondence in complicated ImageNet dataset.

\textbf{Class-conditional Image Generation Fine-tuning.}
Immiscible diffusion can also benefit the fine-tuning. We confirm this intuition with a class-conditional image generation fine-tuning experiment, which use ImageNet to fine-tunes Stable Diffusion which is pre-trained on LAION \cite{schuhmann2022laion} by \cite{stable_diffusion_paper}. Results in Figure \ref{fig:cond_fid} show significant and consistent performance enhancements of immiscible diffusion compared to vanilla baselines. Note that our class-conditional generation uses class names as prompts instead of the class number, so class-conditional fine-tuning and conditional pre-training don't conflict in the form of conditions.

\begin{table}
    \centering
    \small
    \caption{FID evaluations of vanilla and Immiscible Diffusion in Image-to-image Tasks.}
    \vspace{-9pt}
    \resizebox{0.40\linewidth}{!}{
    \begin{tabular}{c|cc}
        \Xhline{2\arrayrulewidth}
        Models & Vanilla SD & \makecell{Immiscible SD \\ KNN} \\
        \hline
        In-painting & 18.35 & \textbf{17.32} \\
        Out-painting & 29.34 & \textbf{27.57} \\
        \Xhline{2\arrayrulewidth}
    \end{tabular}
    }
    \label{tab:i2i}
    \vspace{-3pt}
\end{table}

\textbf{Free-prompt Conditional Image Generation from Scratch.}
To test immiscible diffusion's effects on free prompts other than limited classes of prompts, we train Stable Diffusion~\cite{stable_diffusion_paper} from scratch on MSCOCO~\cite{mscoco}. Results also suggest a performance boost with immiscible diffusion.

\vspace{-6pt}
\subsection{Image Editing}
\vspace{-6pt}

We perform in-painting and out-painting, two classic image-to-image tasks, with vanilla and immiscible diffusion respectively. Both models are Stable Diffusion (SD) models trained on ImageNet dataset. For the in-painting, we load images from ImageNet, replacing its center with Gaussian noise, and we let the model to repair the missing parts of the images. The out-painting task is similar, but everything other than the center part is replaced with Gaussian noise. We compare the completed images with the ImageNet dataset to calculate the FID, as shown in Table \ref{tab:i2i}. We see that in both tasks the immiscible models exhibit better completed images. We further qualitatively provide a few comparisons in Figure \ref{fig:i2i}, where immiscible diffusion enjoys a better overall output which we ascribe to its better preserving of source information.

\vspace{-12pt}
\subsection{Robotics Planning}
\label{sec:robotics}

\vspace{-6pt}

\begin{figure*}[h]
  \vspace{-6pt}
  \centering
  \includegraphics[width=0.80\textwidth]{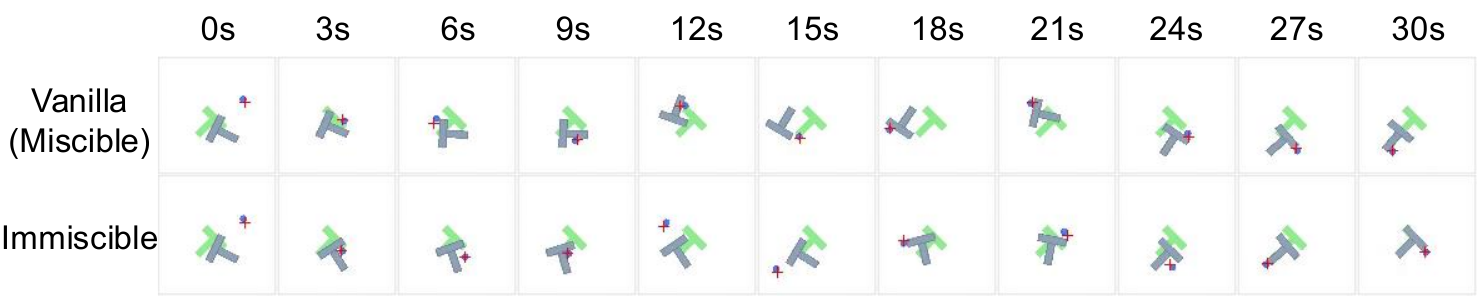}
  \vspace{-6pt}
  \caption{\textbf{Immiscible diffusion boosts the performance of diffusion policy.} The robot (circle) needs to push the T-shaped object (gray) into the desired place (green).}
  \label{fig:diffusion_policy}
  \vspace{-24pt}
\end{figure*}

\begin{table}
    \centering
    \footnotesize
    \caption{Immiscible diffusion in robotics planning.}
    \vspace{-12pt}
    \resizebox{!}{0.070\linewidth}{
    \begin{tabular}{c|cc}
        \Xhline{\arrayrulewidth}
        Experiment & \makecell{Vanilla \\ (Miscible)} & \makecell{Immiscible \\ KNN k=2} \\
        \hline
        \makecell{Ave. Coverage for \\ Last 10 Ckpts (\%)} & 79.56\% & 82.83\% \\
        \hline
        Max Coverage (\%)  & 85.71\% & 86.74\% \\
        \Xhline{\arrayrulewidth}
    \end{tabular}
    }
    \label{tab:diffusion_policy}
    \vspace{-24pt}
\end{table}

To further demonstrate the generalization ability of immiscible diffusion, we apply it onto PushT, a task in diffusion policy \cite{diffusion_policy} for robotics, which let the robot to push a T-shaped object to a desired place. Our experiments are performed on the simulated PushT task explained in Figure \ref{fig:diffusion_policy}, and with the data provided in \cite{diffusion_policy}. Each experiment is trained for $3,000$ epochs with $3$ seeds, where the averages are taken as the results. We take the average area coverage by the T-shaped object onto the desired destination as the metric \cite{diffusion_policy}. 

The experiment results are shown in Table \ref{tab:diffusion_policy}. We see that area coverage is significantly improved with immiscible diffusion. To illustrate this improvement more directly, we show a typical improvement case in Figure \ref{fig:diffusion_policy}. We observe a more accurate pushing process without errors for immiscible diffusion policy where the vanilla one fails due to errors during the pushing.

\subsection{Immiscible Diffusion Beyond Image-noise Correlations}
\label{sec:results_factoring}

\begin{figure}
    \vspace{-18pt}
    \centering
    \includegraphics[width=0.9\linewidth]{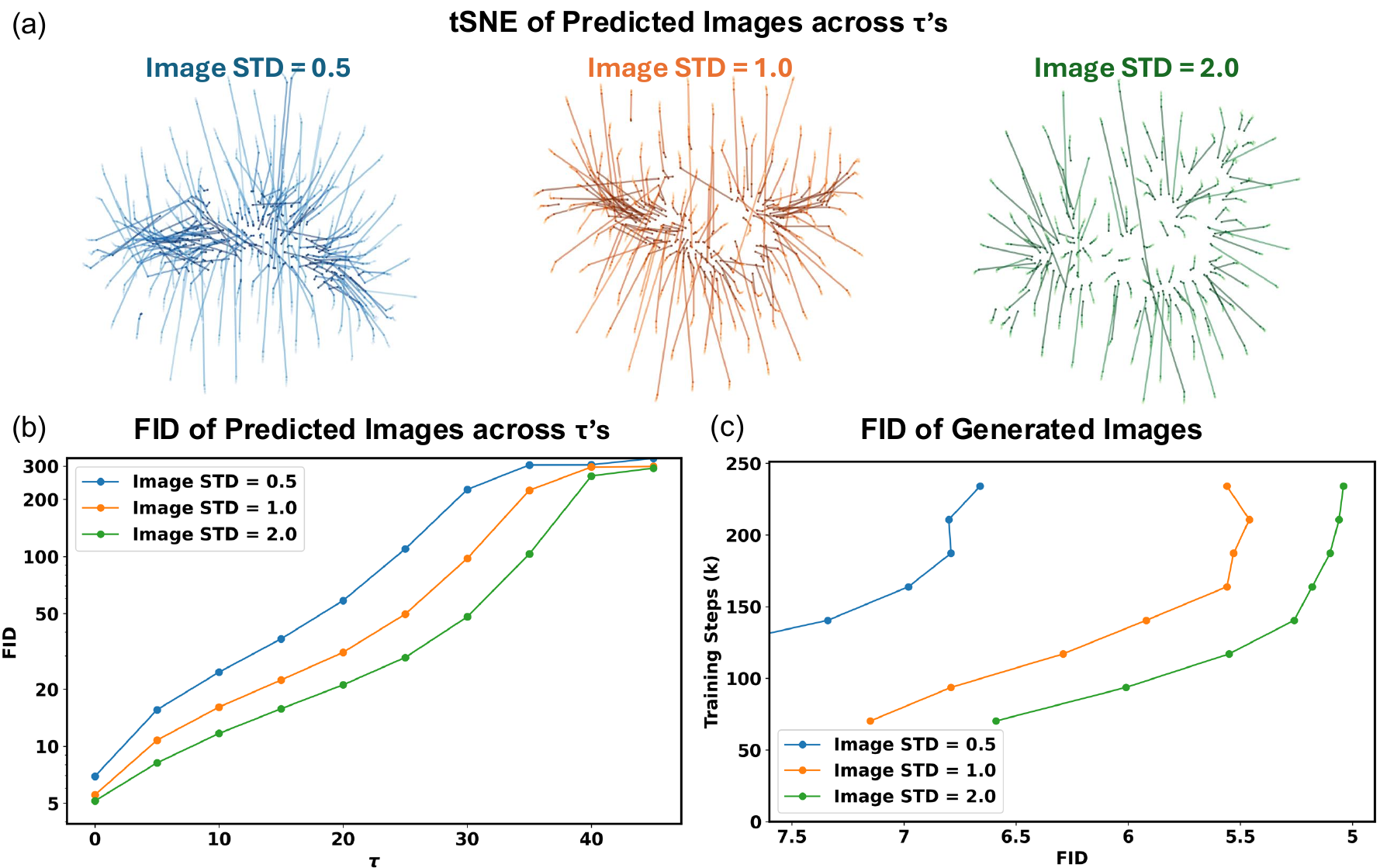}
    \vspace{-9pt}
    \caption{\textbf{Analysis on DDIM with images normed to different pixel value STDs.}}
    \label{fig:factoring}
    \vspace{-18pt}
\end{figure}

As discussed in Section \ref{sec:methods_factoring}, image scaling can intuitively achieve immiscible diffusion without enforcing image-noise correlations. Indeed, in Figure \ref{fig:factoring}, we compare scaling pixel values of normed images to different STD's before performing diffusion with DDIM, whose default is to norm image to a pixel value $STD = 0.5$. Indeed, Figure \ref{fig:factoring} (a) shows that larger STD helps to reduce the confusion in denoising caused by miscible diffusion. Consequently in Figure \ref{fig:factoring} (b), we observe that experiments with larger image STD enjoys lower FIDs of predicted images during denoising, and finally in Figure \ref{fig:factoring} performs better with faster training and better convergence performance. These results are particularly interesting as the scaling can increase the SNR of the final diffusion step, which was shown harmful to the diffusion \cite{Lin_2024_WACV}, further suggesting immiscible diffusion's strong boosts to the diffusion models. In addition, these results show that immiscible diffusion is not necessarily bounded with image-noise correlations.

\vspace{-12pt}
\section{Discussions}
\vspace{-6pt}

In this section, we discuss the distinction of immiscible diffusion to some seemingly similar methods, and compare our immiscible diffusion implementations to highlight their respective advantages.

\textbf{What is the difference between immiscible diffusion, flow matching~\cite{flow_matching} and rectified flow~\cite{rect_flow}?}
We compare immiscible diffusion with flow matching \cite{flow_matching} and vanilla DDIM~\cite{DDIM} in Figure \ref{fig:comparison_fm}. Compared to DDIM, flow matching linearizes the diffusion trajectory. However, its diffusion trajectories (flows) can still arrive at the same noise point from different images, so flow matching is still miscible. Immiscible flow matching aims to reduce such miscibility to let each diffusion trajectory mix less, to ease the denoising. Rectified flow~\cite{rect_flow} can also make \textit{denoising} paths distinct. However, in their method, each image is \textit{diffused} to all the noise space, so their diffusion is still miscible.

\textbf{What is the relation between immiscible diffusion and batch-wise OT?}
Batch-wise OT (linear assignment) can be \textit{one} of the immiscible diffusion implementations. It preferably assign noise to closer images, so images would not be diffused to the whole noise space. Therefore, the mixing of diffusion trajectories from different images reduces, and the diffusion is more immiscible. Step-by-step feature analysis in Section \ref{sec:immiscible_denoising} clearly demonstrate how the linear assignment makes diffusion immiscible and boost the diffusion's performance. At the same time, batch-wise OT only reduces the image-noise average distance by $\sim2\%$~\cite{immiscible_diffusion_nips} and the denoising STD by $\sim4\%$~\cite{Multisample_Flow_Matching}. Therefore, we attribute batch-wise OT's training accelerations mainly to realizations of immiscible diffusion.

\textbf{Which implementation should I use, batch-wise Linear assignment or KNN?}
While there are numerous ways to achieve immiscible diffusion, here we compare the implementations we take for the reader's reference. Since the setting of image scaling depends heavily on the image normalization method taken by the baseline diffusion models, we focus on comparison between batch-wise linear assignment and KNN. While both methods help to enforce the image-noise correlation during diffusion, batch-wise linear assignment keeps the noise space strictly Gaussian, and achieve better immiscibility in the noise space. The average $L_2$ distance between noise clusters assigned to each image for batch-wise linear assignment is $4.11 \pm 0.37$, which is higher than that of KNN, which is $2.17 \pm 0.35$. Therefore, as shown in Figure \ref{fig:feature_analysis} (d), it achieves better FID in noisy layers than KNN. However, batch-wise linear assignment suffers from computational cost of $O(n^3)$, which is higher than the KNN's $O(n)$. Furthermore, in Figure \ref{fig:tSNE_noise_compare}, we show the noise points assigned to the same image using batch-wise assignment and KNN. We observe that the KNN noise points distribute in a more continuous manner, which we posit to contribute to KNN's better performance in denoising un-noisy layers, as also indicated in Figure \ref{fig:feature_analysis} (d). 

\textbf{What $k$ should I use for KNN Immiscible Diffusion?}

 We offer detailed setting of $k's$ in Tab.~\ref{tab:best_k}, where we find the best $k$ generally increases with the rise of diffusion dimensions. We find that while the best $k$ for the same dataset's diffusion dimension is generally the same, with small fluctuations observed, datasets with larger data dimension needs larger $k$ to provide stronger immiscibility. It is noteworthy that for each experiment, we only try $k = 1,2,4,8,16,32,64,128,...$ to save computational resources. Finer experiments with more $k$'s will provide more precise results. Nevertheless, in Tab.~\ref{tab:k_impact}, we find the improvements can persist in a wide spectrum of $k$.

\begin{table*}
    \centering
    \caption{Best $k's$ in KNN Immiscible Diffusion Implementation.}
    \begin{tabular}{c|ccccc}
        \Xhline{2\arrayrulewidth}
        Model & DDIM & Flow Matching & Consistency Model & Stable Diffusion\\
        Dataset & CIFAR-10 & CIFAR-10 & CIFAR-10 & ImageNet-1k \\
        \hline
        Diffusion Dimension & 3,072 & 3,072 & 3,072 & 4,096 \\
        Best $k$ & 8 & 4 & 4 & 64 \\
        $L_2$ Dist. $\Delta$ (\%) & -1.58\% & -1.10\% & -1.10\% & -2.32\%\\
        \Xhline{2\arrayrulewidth}
    \end{tabular}
    \label{tab:best_k}
\end{table*}

\begin{table}[htbp]
\vspace{-9pt}
\caption{Impact of $k$ on Best FID.}
\vspace{-9pt}
\centering
\resizebox{0.55\linewidth}{!}
{
    \begin{tabular}{c|cccccc}
        \Xhline{\arrayrulewidth}
        $k$ & Baseline & 2 & 4 & 8 & 16 & 32 \\
        \midrule
        Best FID & 6.63 & 4.94 & 4.77 & 4.45 & 4.85 & 5.19 \\
        \Xhline{\arrayrulewidth}
    \end{tabular}
}
\label{tab:k_impact}
\vspace{-9pt}
\end{table}

\vspace{-12pt}

\section{Conclusion}
\label{sec:conclusion}

In this work, we systematically revisit immiscible diffusion, a physics-inspired method aiming to boost diffusion training efficiency. Our experiments firstly show that image-noise correlation introduced by immiscible diffusion empirically does not alter the diversity of generated images due to the intrinsic image-noise correlation in vanilla diffusion models. Detailed feature analysis shows how immiscible diffusion step-by-step enhances the denoising outputs. Based on these findings, we improve the immiscible diffusion concept, which does not require doing image-noise pairing nor even image-noise correlations. We offer a few new immiscible diffusion implementations, achieving training efficiency boosts up to $>4\times$, across diverse baseline diffusion models and on various image generation tasks, image editing and robotics planning. Our method points out the diffusion trajectory miscibility problem, a generally existing problem in diffusion training dragging its efficiency, which could be a fundamental direction to explore towards high-efficiency diffusion training.

\bibliographystyle{splncs04}
\bibliography{main}

\newpage
\clearpage
\setcounter{page}{1}

\section{Supplemental Materials}
\subsection{Experiment Setup Details}
\label{sec:spp_exp_setup}

\begin{table}
\caption{Image Generation Experiment setting.}
\vspace{-3pt}
\centering
\resizebox{1.0\linewidth}{!}{
    \begin{tabular}{c|ccccccc}
        \Xhline{2\arrayrulewidth}
        Model & \makecell{Consistency\\Model} & DDIM & \makecell{Flow\\Matching} & \makecell{Stable Diffusion\\Class-conditional} & \makecell{Stable Diffusion\\Conditional} & \makecell{Stable Diffusion\\Fine-tuning}\\
        \midrule
        Dataset & CIFAR-10 & CIFAR-10 & CIFAR-10 & ImageNet & MS-COCO & ImageNet \\
        Batch Size & 512 & 256 & 256 & 2048 & 2048 & 512\\
        Resolution & $32 \times 32$ & $32 \times 32$ & $32 \times 32$ & $256 \times 256$ & $256 \times 256$ & $256 \times 256$ \\
        Devices & $4 \times A6000$ & $1 \times A5000$ & $1 \times A6000$ & \makecell{$8 \times A800$ or \\ $4 \times A6000$} & $4 \times A6000$ & $4 \times A6000$ \\
        \Xhline{2\arrayrulewidth}
    \end{tabular}
}
\label{tab:train_param}
\end{table}

The training hyperparameter are listed in Table \ref{tab:train_param}. Unspecified hyper-parameters are taken the same as those in their baseline methods' original papers. For evaluations, we compare the generated images by Immiscible Diffusion and the baseline using the quantitative evaluation metric FID \cite{FID}. Note that for Consistency Models, we use the single-step generation consistency training. For DDIM, we add no noise during the sampling and use linear scheduling to select sampling steps. For Stable Diffusion, we directly use the implementation from Diffusers of Huggingface team \cite{stable_diffusion_paper}. For fine-tuning, we use Stable Diffusion v1.4 \cite{stable_diffusion_paper} as the pre-trained model. Image in-painting and out-painting does not involve additional training, and details on robotics plannings will be listed in the Section \ref{sec:robotics}.

\subsection{Qualitative Comparisons of Image-to-image Tasks}
\begin{figure}
  \centering
  \includegraphics[width=\textwidth]{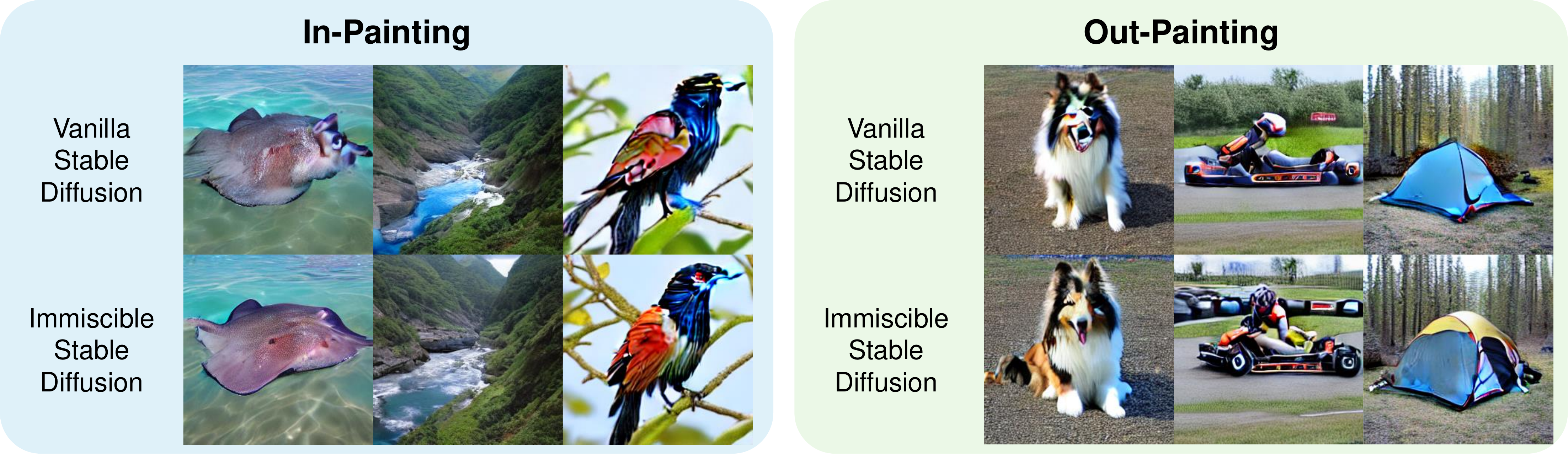}
  \caption{\textbf{Qualitative Illustration of Immiscible Diffusion in Image-to-image Tasks.} We notice that immiscible diffusion can better preserve existed information (i.e. edge information in in-painting and center information in out-painting) so as to provide better overall completed images.}
  \label{fig:i2i}
\end{figure}

We provide qualitative comparisons of image-to-image tasks in Fig.\ref{fig:i2i}, as discussed in the main sections.

\subsection{tSNE of Noise Point Clusters for Different Methods}

\begin{figure*}
  \centering
  \includegraphics[width=0.55\linewidth]{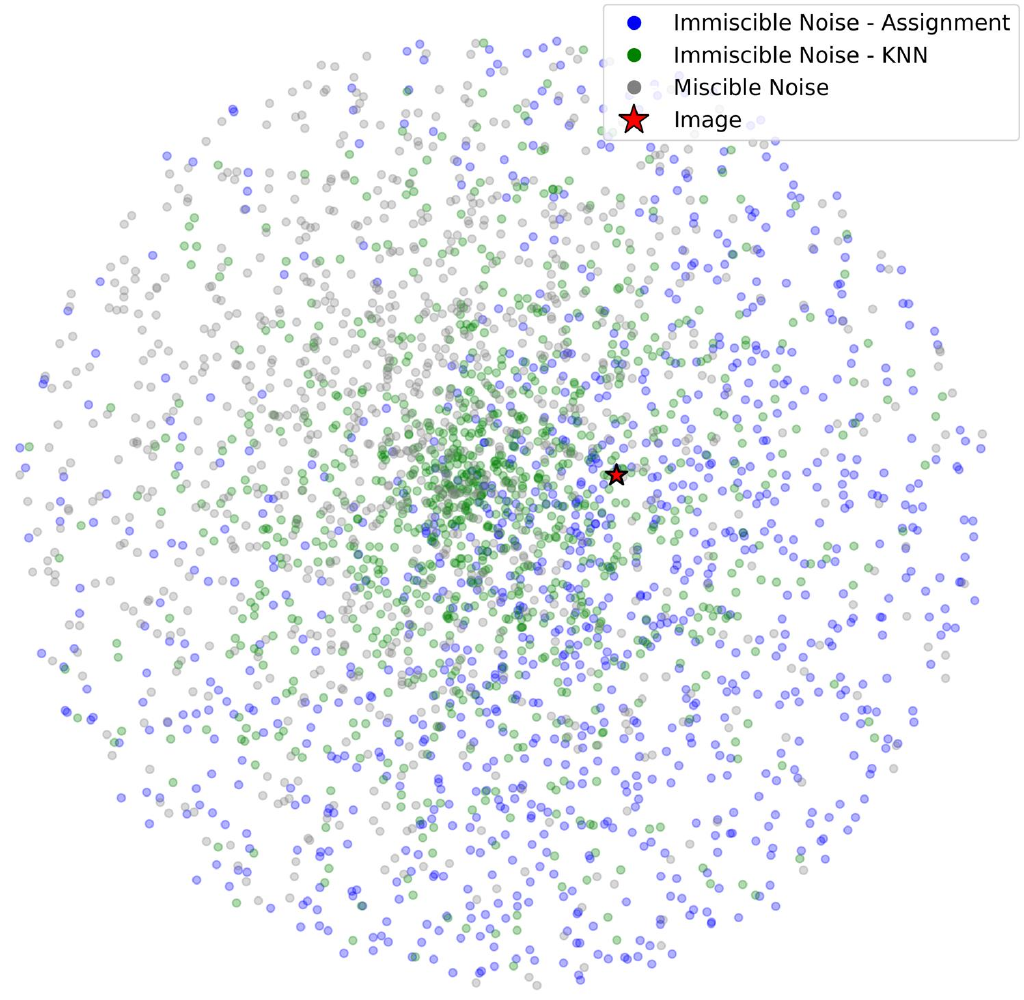}
  \caption{\textbf{tSNE of noise clusters belonging respectively to vanilla, assign and KNN DDIM.} }
  \label{fig:tSNE_noise_compare}
\end{figure*}

The tSNE figures of noise point clusters for different methods are offered in Fig. \ref{fig:tSNE_noise_compare}.

\subsection{Illustrations Comparing Immiscible Diffusion to Other Common Methods}

\begin{figure*}
  \centering
  \includegraphics[width=0.90\linewidth]{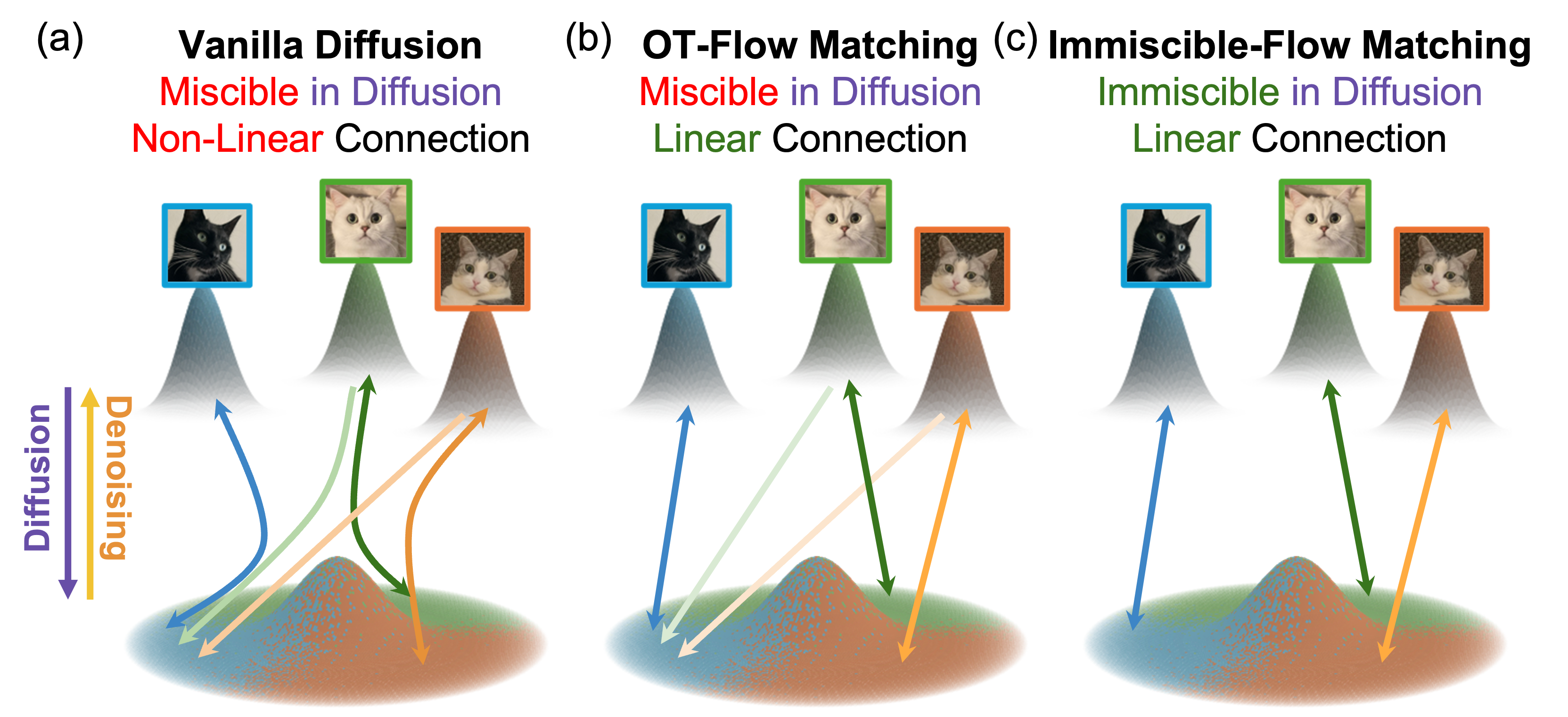}
  \caption{Comparison between Diffusion Models, Flow Matching and Immiscible Flow Matching.}
  \label{fig:comparison_fm}
\end{figure*}

We illustrate the comparison between immiscible diffusion and other common methods in Fig.\ref{fig:comparison_fm}

\end{document}